\documentclass[11pt,a4paper]{llncs}
\usepackage{amsmath}
\usepackage{amssymb}
\usepackage{graphicx}
\usepackage{marvosym}
\usepackage{url}
\usepackage{fancyhdr}

\usepackage{algorithm}
\usepackage{algpseudocode}
\usepackage[colorlinks=true,linkcolor=blue,citecolor=blue,urlcolor=blue]{hyperref}

\usepackage{geometry}
\newcommand{\keywords}[1]{\par\addvspace\baselineskip
\noindent\keywordname\enspace\ignorespaces#1}

\fancypagestyle{firstpage}{\fancyhf{}
}

\begin{document}


\title{\LARGE{Household Movement Detection in Mixed-Format Occupancy Data Using LLM-Based Entity Resolution}}


%
%
\author{\large{Sasirekha Oguri  \and John R. Talburt \and Mert Can Cakmak}}
\institute{\large{Center for Entity Resolution and Information
Quality (ERIQ) \\ University of Arkansas - Little Rock}}

%


%
%


\maketitle

\thispagestyle{firstpage}

\begin{abstract}
Entity resolution (ER) typically relies on pairwise similarity comparisons between records, which limits its ability to capture indirect relationships present in demographic occupancy data. An important indirect pattern arises from household movement, where multiple individuals relocate together across addresses, but detecting such patterns is difficult due to mixed-format records, noise, duplication, and the absence of stable identifiers. This paper proposes an AI-enhanced framework for detecting indirect entity links associated with household movement in unstandardized name--address data. The approach integrates prompt-based large language model (LLM) named entity recognition for extracting personal names and addresses without extensive preprocessing, semantic text embeddings for robust similarity computation, and graph-based reasoning to infer group-level movement patterns. Experimental evaluation on SPX benchmark datasets (S8--S12) generated using the Synthetic Occupancy Generator demonstrates that incorporating indirect household movement evidence improves recall by 8--15\% while maintaining high precision, yielding F1-score gains of 6--8\% over a strong pairwise baseline.

\end{abstract}

\section{Introduction}

Entity resolution (ER) is a foundational task in data integration, concerned with
identifying and linking records that refer to the same real-world entity across
disparate, heterogeneous, and often noisy data sources \cite{qian2017active,fellegi1969theory,christen2012data}. ER underpins a wide range of applications,
including customer data unification, population analytics, fraud detection,
healthcare records integration, and government service delivery. As enterprise
systems increasingly incorporate large language models into operational data
pipelines, the ability of these models to preserve semantic and syntactic
integrity in structured contexts has become an important consideration
\cite{maclean2026evaluating}.

Most ER systems operate under a pairwise matching paradigm, in which linking
decisions are made by comparing individual record pairs based on attribute
similarity. While effective in many structured settings, pairwise ER is limited
in its ability to capture \emph{indirect relationships} that arise from contextual
or group-based patterns. In practice, valid links may exist between records that
do not directly match but are connected through shared associations or collective
behavior \cite{bhattacharya2007collective,getoor2012er}.

One particularly important class of indirect linking patterns arises from
\emph{group membership}. When entities participate in stable or semi-stable
groups—such as households, families, or organizational units—relationships can
be inferred through co-occurrence and shared transitions \cite{ebeid2022graph}.
Household movement in occupancy data (name and address records) exemplifies
this phenomenon. A household movement pattern occurs when two or more individuals
who previously co-resided at one address are later observed at a different
address, implying a coordinated relocation of the group. Such patterns provide
powerful evidence for linking individuals and reconstructing household histories,
even when individual records are noisy or inconsistent.

Detecting household movement, however, is substantially more complex than tracking
individual address changes. Real-world occupancy data frequently exhibits
incomplete temporal information, duplication, spelling variation, mixed-format
records, and missing attributes. Household transitions are often asynchronous:
members may appear at different times, some may drop out of the data entirely,
and others may change names due to marriage or cultural practices. These factors
render rigid string matching and purely temporal approaches brittle and prone to
false negatives \cite{binette2022almost}. Recent research on LLM-based name and
address parsing systems demonstrates the feasibility of extracting structured
entities directly from heterogeneous reference formats \cite{tarannum2026system},
highlighting the potential of prompt-based semantic extraction in mixed-format
occupancy data.

Recent work has explored graph-based ER techniques to leverage transitive and
collective evidence for identifying entity profiles and movement patterns
\cite{ebeid2022graph}. Other studies have demonstrated the potential of large
language models (LLMs) to extract structure from unstructured or mixed-format
references \cite{devlin2019bert,li2020ditto}. At the same time, policy-aware and
governance-oriented generative AI frameworks emphasize the importance of
controlled, auditable model deployment when applied to sensitive demographic
data environments \cite{al2025policy}. While promising, many LLM-based
approaches either require extensive preprocessing, operate on small manually
curated subsets, or lack a scalable mechanism for systematically identifying
candidate regions of interest within large datasets.

This paper addresses these challenges by reframing household movement detection
as a \emph{group-level entity resolution problem}. Rather than focusing solely on
directly matching individuals, the objective is to infer household transitions
through shared membership and semantic similarity, even when names and addresses
are inconsistently recorded. The proposed framework integrates LLM-based named
entity recognition, semantic text embeddings, and graph-based transitive reasoning
to enable scalable detection of both direct and indirect household movement links
in large, heterogeneous datasets.

\section{Related Work}

Household movement detection intersects three major research areas:
(i) classical record linkage and entity resolution (ER), (ii) collective and
graph-based ER, and (iii) representation learning and large language
models (LLMs) for robust matching under noise. Our work contributes at the
intersection of these areas by treating household movement as a \emph{group-level}
inference problem and using semantic representations plus transitive graph
reasoning to recover indirect links that are systematically missed by
pairwise-only methods.

\subsection{Classical Record Linkage and Rule-Based Approaches}

Early record linkage methods formalized matching as a probabilistic decision
problem, often using field-level agreement patterns and thresholds
(e.g., Fellegi--Sunter) \cite{fellegi1969theory}. Practical ER systems frequently
combine such scoring with string similarity (e.g., edit distance \cite{levenshtein1966},
Jaro--Winkler \cite{winkler1990string}) and blocking strategies to scale
to large datasets \cite{christen2012data,sarawagi2002duplicate}. While effective
in structured environments, these methods typically assume stable schemas and
clean, consistently tokenized fields. In real administrative data, however,
mixed-format records, missing fields, abbreviations, and typographical errors
can cause brittle behavior and substantial recall loss.

Address-centric movement detection has also been explored by tracking sequential
address transitions over time. Such approaches can generate false positives in
high-density residences or institutional settings and become unreliable when
timestamps are missing or inconsistent. In addition, household movement is not
always a synchronized, clean transition: members may appear at different times,
disappear, or change identifiers (e.g., through surname changes), making movement
detection harder than individual address-change tracking.

\subsection{Graph-Based and Collective Entity Resolution}

A substantial body of work models ER as a graph problem, where nodes represent
records or entities and edges represent similarity or relational evidence.
Graph-based ER can exploit transitivity and collective evidence to improve recall
over purely pairwise matching \cite{bhattacharya2007collective,getoor2012er}.
Clustering-oriented frameworks and deduplication workflows address ambiguous,
multi-reference settings by iteratively merging partial views and reconciling
conflicting evidence \cite{benjelloun2009swoosh}. More recent ER research has
also emphasized efficient candidate generation and pruning (e.g., meta-blocking)
to reduce comparisons while preserving recall \cite{papadakis2016metablocking}.

Within this line, Ebeid \cite{ebeid2022graph} describes a rule-based process for
mixed-format references: references are parsed and standardized; addresses are
clustered; and a graph is constructed linking names to address nodes, which is
then traversed to detect household movement patterns. While effective, such approaches
still depend on standardization or deterministic parsing steps that can be
fragile for highly heterogeneous records.

Our work differs in two important ways: (i) it minimizes reliance on
standardization by using LLM-based extraction and semantic embeddings, and (ii)
it introduces a scalable mechanism (analytical segmentation) to identify
candidate subsets likely to contain household movement patterns, enabling
systematic processing of entire datasets rather than manually curated subsets.

\subsection{Representation Learning and Deep Learning for ER}

In the last decade, ER research increasingly adopted learned representations to
handle noisy, diverse text. Deep learning approaches have shown strong results
on product/entity matching by learning similarity directly from raw attributes
\cite{mudgal2018deep}. Transformer-based models further improved robustness by
capturing context across fields and learning fine-grained token interactions
\cite{devlin2019bert}. Systems such as Magellan provide end-to-end ER workflows,
including feature engineering, labeling, and learning-based matching
\cite{konda2016magellan}.

More recently, pretrained language models have been used directly for ER and
data integration, achieving strong performance without heavy manual feature
engineering (e.g., Ditto) \cite{li2020ditto}. Sentence embedding approaches also
enable scalable similarity computation under noise, particularly when exact
string agreement fails \cite{reimers2019sentence}. Recent work has further explored the use of large language models and
multi-agent architectures for entity resolution. For example, Althaf et al.
\cite{althaf2025multi} propose a multi-agent retrieval-augmented generation
(RAG) framework in which specialized agents coordinate to improve ER
performance beyond single-LLM pipelines. Such approaches demonstrate the
growing interest in leveraging LLM reasoning and modular architectures for
complex linkage tasks. However, most existing LLM-based ER systems focus
primarily on pairwise matching or entity profile reconstruction, whereas the
present work emphasizes scalable detection of indirect, group-level household
movement patterns through embedding-based segmentation and graph reasoning.

Our method aligns with this direction by using semantic embeddings for names and
addresses, but extends beyond typical ER objectives (pairwise matching) by
explicitly inferring household movement via group-level co-occurrence evidence
and transitive link construction.

\subsection{Household Mobility Data and Privacy Constraints}

Real household mobility studies often rely on high-quality administrative
registries or census-linked longitudinal data, but access is heavily constrained
due to privacy and confidentiality concerns. Public research datasets such as
IPUMS provide large-scale demographic resources that are valuable for population
research \cite{ruggles2021ipums}, but they do not directly provide the kind of
noisy, mixed-format operational records common in service delivery systems.

Synthetic data generation frameworks address this gap by enabling controlled
evaluation under realistic degradation while preserving privacy. In this paper,
we evaluate on SPX benchmark datasets generated by the Synthetic Occupancy
Generator (SOG), which provides ground truth while simulating real-world
heterogeneity, duplication, and fragmentation.

In summary, prior work establishes strong foundations in rule-based linkage,
probabilistic record linkage, graph-based ER, and learning-driven approaches.
Recent advances in LLM-based architectures \cite{althaf2025multi} and improved
evaluation methodologies for comparing ER outputs \cite{talburt2026case}
highlight the rapid evolution of the field. However, household movement
detection in mixed-format, noisy data remains challenging due to the need for
group-level inference, incomplete evidence, and scalable processing. The
approach proposed here is distinct in its combination of (i) LLM-based semantic
extraction to reduce dependence on schema normalization, (ii) embedding-based
blocking via analytical segmentation to enable automated large-scale
processing, and (iii) graph-based transitive reasoning to recover indirect
household links that improve recall while maintaining high precision.

\section{Problem Definition}

Detecting household movement in large-scale, real-world occupancy data presents challenges that extend well beyond traditional record linkage or address matching. Unlike curated registries or survey datasets, administrative and operational data rarely provide explicit household identifiers, consistent schemas, or reliable temporal annotations. Instead, household structure and movement must be inferred indirectly from noisy, heterogeneous observations of individuals appearing at different locations over time.

Formally, the problem addressed in this work is the identification of household-level movement events from collections of unstandardized name--address records, where (i) individual records may be duplicated or corrupted, (ii) households are implicit group entities rather than explicitly labeled objects, and (iii) movement events are not directly observed but must be inferred from patterns of shared membership across addresses. In this setting, traditional pairwise entity resolution methods are insufficient, as they focus on linking individual records rather than detecting coordinated transitions of groups.

Three fundamental challenges define this problem: heterogeneous record layouts, degradation of attribute quality, and the computational complexity of inferring group-level movement patterns at scale.

\subsection{Heterogeneous Record Layouts}

Real-world occupancy datasets---such as customer transaction records, clinical admissions, or administrative registries---often lack a consistent schema and may combine structured and unstructured representations within the same dataset. The same household or individual may appear multiple times with attributes distributed across different fields, concatenated into free-text strings, or recorded under different layouts depending on data source, system version, or entry protocol.

As a result, logically equivalent records may differ substantially in surface form. For example, an individual’s name and address may appear as a single unstructured string in one record, while being decomposed into multiple structured fields in another. This heterogeneity complicates downstream processing, as linkage systems must first identify and extract comparable semantic units before any meaningful similarity assessment can occur.

Table~\ref{tab:formats} illustrates this challenge using examples from the S12PX test dataset, which is designed to replicate real-world administrative and census data conditions, including heterogeneous schemas, missing fields, value-level inconsistencies, and duplicate entries.

\begin{table}[ht]
\centering
\caption{Example of heterogeneous record layouts referring to the same individual and address}
\label{tab:formats}
\begin{tabular}{|l|l|l|l|l|}
\hline
\textbf{RecID} & \textbf{Name} & \textbf{Address} & \textbf{City} & \textbf{State Zip} \\
\hline
A960815 & TERRIE D LOPZE & r2623 fiddlestick crcle & lutz & fl u33559 \\
\hline
\end{tabular}

\vspace{0.3cm}

\begin{tabular}{|l|l|l|l|l|l|l|}
\hline
\textbf{RecID} & \textbf{First} & \textbf{Last} & \textbf{Street \#} & \textbf{Street} & \textbf{City} & \textbf{State Zip} \\
\hline
B982950 & TERRIE D & ZLOPEZ & 2623 & fiddlestick cir & LUTZ & FL 33559 \\
B972532 & TERRIE D & LOPZE & 2623 & FIDDLYESTICK CIR & Lutz & fl 33559 \\
\hline
\end{tabular}
\end{table}

Such inconsistencies prevent direct field-level comparison and motivate approaches capable of extracting semantic structure from mixed-format inputs without relying on extensive preprocessing or schema normalization.

\subsection{Data Quality Degradation}

Beyond structural heterogeneity, demographic and occupancy data are subject to significant degradation at the attribute level. Names and addresses commonly exhibit typographical errors (e.g., \emph{crcle} vs.\ \emph{cir}), phonetic variations (e.g., \emph{Fiddlestick} vs.\ \emph{Fiddlyestick}), abbreviations (e.g., \emph{St} vs.\ \emph{Street}), missing components (such as apartment numbers), outdated values, and data-entry artifacts including extraneous characters or annotations.

These distortions substantially reduce the effectiveness of exact or rule-based matching strategies and increase both false negatives and false positives in linkage tasks. Robust household movement detection therefore requires methods that can tolerate partial matches, semantic variation, and noisy representations while preserving discriminative power.

\subsection{Household Movement Detection at Scale}

Household movement detection differs fundamentally from tracking individual address changes. A household is a group entity whose membership and location may evolve over time in complex and asynchronous ways. Members of the same household may appear to move at different times, some may temporarily disappear from the data, and new individuals may join mid-transition. Consequently, household movements rarely manifest as clean, synchronized transitions across all members.

In large datasets spanning thousands of addresses and extended time horizons, these dynamics result in fragmented and incomplete movement evidence. A household may appear to relocate from Address~A to Address~B and later to Address~C, but only through partial overlaps of membership at each stage. Formatting differences, name changes, and missing records can further obscure continuity, rendering direct pairwise comparisons insufficient.

Compounding this challenge, movement events are typically unlabeled: there is no explicit indicator specifying when or whether a household has moved. Instead, movement must be inferred indirectly from recurring patterns of shared individuals across distinct addresses. As dataset size grows, the number of possible groupings and transitions increases combinatorially, making exhaustive comparison computationally infeasible.

These characteristics define household movement detection as a group-level inference problem under uncertainty, requiring approaches that can integrate noisy evidence, reason over indirect connections, and scale efficiently to large, heterogeneous datasets.

\section{Proposed Method: NER--Embedding--Graph Framework}

To address the challenges outlined in Section~3, we propose a modular framework that integrates large language model (LLM)--based named entity recognition, semantic text embeddings, and graph-based reasoning to detect household movement in noisy, mixed-format occupancy data. The method is explicitly designed to operate under conditions where schema consistency, attribute standardization, and explicit household identifiers are unavailable.

Rather than relying on rigid preprocessing or deterministic rules, the proposed framework incrementally refines unstructured records into increasingly structured representations. Raw records are first decomposed into semantic entities, then grouped into analytically meaningful subsets, and finally analyzed at the household level to infer movement events through shared membership patterns. This layered design enables robust group-level inference while remaining scalable to large datasets.

For completeness and reproducibility, a detailed step-by-step operational description of the pipeline is provided in Appendix~\ref{appendix:pipeline}.

\subsection{Pipeline Overview}

Given a dataset of occupancy records containing unstructured or semi-structured name and address information, the proposed method proceeds through six conceptual stages:
(i) extraction of personal names and addresses using LLM-based named entity recognition,
(ii) semantic embedding of extracted entities,
(iii) construction of similarity graphs to identify analytically relevant record subsets,
(iv) household identification within these subsets,
(v) consolidation of duplicate records through leader--follower mapping, and
(vi) inference of household movement events and indirect links.

This decomposition reflects the observation that household movement is not directly observable at the record level but must instead be inferred through intermediate relational structures. Algorithm~\ref{alg:pipeline} summarizes the overall processing flow.

\begin{algorithm}[ht]
\caption{NER--Embedding--Graph Household Movement Detection}
\label{alg:pipeline}
\begin{algorithmic}[1]
\State Extract names using LLM-based NER
\State Isolate and normalize addresses
\State Embed names and addresses
\State Build similarity graph and extract analytical segments
\State Form households using address similarity
\State Detect duplicates and construct leader--follower mappings
\State Generate pre-move candidate sets
\State Detect household movements and indirect links
\State Output linkage and movement records
\end{algorithmic}
\end{algorithm}

\subsection{LLM-Based Named Entity Recognition}

The first stage extracts personal names from mixed-format occupancy records using the instruction-tuned large language model GEMMA-2-2B-IT~\cite{gemma2_2b_it}. In contrast to traditional token-level or rule-based NER systems, which assume clean field boundaries and consistent formatting, this approach treats each record as a free-text sequence and relies on semantic understanding to isolate person names.

All textual fields associated with a record are concatenated into a single input string. A few-shot prompt instructs the model to extract only personal name entities while explicitly ignoring numeric tokens, street information, city names, PO Boxes, and ZIP codes. This formulation allows the model to generalize across heterogeneous layouts, incomplete delimiters, and embedded name--address combinations. Representative examples are shown in Table~\ref{tab:ner_examples}.

\begin{table}[ht]
\centering
\caption{Examples of LLM-based name extraction from mixed-format occupancy records}
\label{tab:ner_examples}
\begin{tabular}{|p{6.5cm}|p{4.5cm}|}
\hline
\textbf{Input Record (Concatenated Fields)} & \textbf{Extracted Name} \\
\hline
terrie d, zlopez, 2623 fiddlestick cir, lutz, fl & terrie d, zlopez \\
\hline
z, cooling, 2406 lake road, elot 15, belton & z, cooling \\
\hline
\end{tabular}
\end{table}

Inference is performed using deterministic decoding parameters (temperature = 0.01, top-$k$ = 1) to ensure reproducibility and to minimize output variability. After name extraction, the address component is isolated by removing the extracted name substring from the original input. Light normalization steps, such as whitespace collapse and punctuation trimming, are applied to produce a comparable address representation without enforcing full standardization.

On the S12PX dataset (6,000 records), this process achieved a name extraction accuracy of 97.38\%, demonstrating robustness to spelling errors, abbreviations, missing delimiters, and mixed-case formatting.

\subsection{Semantic Embedding of Names and Addresses}

Extracted names and addresses are independently transformed into dense vector representations using the BAAI/bge-m3 sentence embedding model. Separating name and address embeddings allows the framework to capture distinct similarity signals: name embeddings emphasize phonetic and lexical variation, while address embeddings capture semantic and structural similarity across location descriptions.

Similarity between records is computed using cosine distance:
\[
\text{Sim}_{name}(i,j) = \cos(\mathbf{E}_{name}^{(i)}, \mathbf{E}_{name}^{(j)}), \quad
\text{Sim}_{addr}(i,j) = \cos(\mathbf{E}_{addr}^{(i)}, \mathbf{E}_{addr}^{(j)}).
\]
This representation enables the framework to tolerate partial matches and noisy values while preserving discriminative power.

\subsection{Similarity Graph Construction and Analytical Segments}

To limit the combinatorial complexity of downstream analysis, records are first organized into analytically meaningful subsets. An undirected similarity graph $G = (V, E)$ is constructed, where each node represents a record. An edge is added between records $i$ and $j$ if either name similarity or address similarity exceeds predefined thresholds:
\[
\text{Sim}_{name}(i,j) \ge \tau_{name} \;\; \text{or} \;\;
\text{Sim}_{addr}(i,j) \ge \tau_{addr}.
\]

The connected components of $G$ define \emph{analytical segments}, which serve as localized neighborhoods of potentially related records. This OR-based edge criterion ensures high recall during segmentation, allowing weak evidence in one attribute (e.g., name variation) to be compensated by stronger evidence in another (e.g., address similarity).

\subsection{Household Formation Within Segments}

Within each analytical segment, households are identified using address similarity alone. Records whose pairwise address similarity exceeds $\tau_{addr}$ are linked, and the resulting connected components are labeled as households. This step reflects the assumption that co-residence is primarily determined by shared or semantically similar addresses, while name similarity alone is insufficient to define household membership.

By performing household formation within analytical segments rather than globally, the framework avoids spurious merges and reduces sensitivity to common addresses or high-frequency names.

\subsection{Duplicate Detection and Leader--Follower Consolidation}

Within each household, duplicate records and alternate name representations for the same individual are identified using name similarity. Records exceeding a duplication threshold $\tau_{dup}$ are grouped into duplicate clusters, while unlinked records are labeled as unique (UNQ).

To prevent duplicate records from artificially inflating household membership and movement evidence, each duplicate cluster is consolidated using a leader--follower strategy. The first record in the cluster is designated as the leader, while all other records are treated as followers. A pre-move candidate set is then constructed for each household by retaining one representative per individual (the leader) along with all UNQ records.

\subsection{Household Movement Detection}

Household movement events are inferred by comparing households pairwise within the same analytical segment. A movement is detected when two households share at least two individuals whose names exceed a similarity threshold $\tau_{move}$. Requiring multiple shared individuals reduces the likelihood of coincidental overlap and provides evidence of coordinated relocation.

Detected movement events are used to construct indirect links connecting recurring individuals across distinct addresses. These indirect links enable transitive, group-level inference that extends beyond traditional pairwise entity resolution and captures household transitions even when records are fragmented or partially observed.

\section{Proof of Concept: Experiments on the SPX Datasets}

This section presents a proof-of-concept evaluation of the proposed
NER--Embedding--Graph framework using synthetic occupancy datasets generated by
the Synthetic Occupancy Generator (SOG). The objective is to demonstrate---using a
fully traceable, end-to-end example---how the framework detects \emph{indirect}
household movement links that are not recoverable using conventional pairwise
entity resolution (ER). 

\subsection{Dataset Overview}

The experiments are conducted on datasets from the SPX benchmark family (S8--S12),
generated using the Synthetic Occupancy Generator (SOG), a scenario-driven synthetic
data framework designed to support ER research by producing realistic, privacy-preserving
residential occupancy histories \cite{talburt2024sog}. We evaluate SOG's \emph{degraded
external view}, where internal identity identifiers are removed and records are disrupted
through duplication, heterogeneous layouts, formatting variation, abbreviation drift,
attribute omission, and typographical noise. This setting mirrors common challenges in
administrative and census-style occupancy data.

SOG generates data using an \emph{internal view of identity}, in which each synthetic
individual is assigned a unique identifier and a complete, temporally ordered occupancy
history (a sequence of name--address observations over time). These histories are produced
from explicit life-event scenarios that simulate realistic population behavior, including
single-person mobility, household formation through co-residence or marriage, surname
changes, separation, and repeated address transitions. In multi-person scenarios, SOG models
periods of shared occupancy and coordinated moves, enabling household-level evaluation.

To produce data suitable for ER evaluation, SOG generates an \emph{external view} by removing
identity identifiers and disrupting the records (duplication, layout shifts, missing values,
typos, abbreviation variation, and value perturbation). As a result, identity continuity and
household structure must be inferred solely from observable attributes such as names and addresses.

Table~\ref{tab:sog_views} summarizes the conceptual distinction between SOG's internal ground-truth
representation and the disrupted external view used by linkage algorithms.

\begin{table}[h]
\centering
\caption{Internal and external views of identity in SOG-generated datasets}
\label{tab:sog_views}
\begin{tabular}{|l|l|}
\hline
\textbf{Internal View (Ground Truth)} & \textbf{External View (Observed Data)} \\
\hline
Unique identity identifiers & Identifiers removed \\
Complete occupancy histories & Fragmented name--address records \\
Explicit temporal ordering & Partial or missing temporal cues \\
Canonical names and addresses & Noisy, duplicated, inconsistent values \\
\hline
\end{tabular}
\end{table}

The SPX benchmark family consists of datasets labeled S8 through S12, representing progressively
more challenging configurations derived from the same SOG framework. As the dataset index increases,
the data exhibits higher levels of corruption, duplication, household fragmentation, and indirect
movement patterns. Table~\ref{tab:spx_overview} provides a high-level characterization.

\begin{table}[h]
\centering
\caption{SPX benchmark datasets used in the experiments}
\label{tab:spx_overview}
\begin{tabular}{|l|l|}
\hline
\textbf{Dataset} & \textbf{Relative Characteristics} \\
\hline
S8  & Lower noise, fewer duplicates, simpler household structures \\
S9  & Moderate duplication and address variation \\
S10 & Increased name and address corruption \\
S11 & Higher household churn and partial membership overlap \\
S12 & Highest complexity with chained household movements \\
\hline
\end{tabular}
\end{table}

Although SOG provides definitive internal ground truth, not all scenarios contain sufficient observable
evidence for reliable linkage in the external view. For example, some household transitions may involve
only one overlapping individual, surname changes can reduce name similarity below practical thresholds,
and explicit timestamps may be absent after disruption, preventing reliable inference of movement direction.
Accordingly, the evaluation focuses on movement patterns that are inferable from observable name and address evidence.

All personally identifiable information (PII) such as Social Security Numbers (SSNs) and Dates of Birth (DOBs)
is excluded from the analysis.

\subsection{Illustrative Input Example}

To provide a transparent end-to-end illustration, we use a fixed subset of 8 records from the S12PX scenario.
The full S12PX dataset contains approximately 6{,}000 synthetic occupancy records and exhibits heterogeneous
and noisy entries in which name and address components may be interwoven, inconsistently formatted, or partially
corrupted---conditions intended to resemble real-world administrative and census data quality challenges.

Table~\ref{tab:raw_input} shows the \emph{raw} multi-field structure of the illustrative records (PII columns omitted),
highlighting mixed-format input (multiple address fragments and PO-box fields). These records are subsequently processed
using LLM-based name extraction and address isolation to produce a structured representation used in all downstream steps.
The same 8 records are used consistently throughout Sections~5.3--5.5 to ensure end-to-end traceability.

\begin{table}[h]
\centering
\caption{Illustrative raw mixed-format input records from S12PX (PII fields omitted)}
\label{tab:raw_input}
\scriptsize
\begin{tabular}{|l|p{0.14\linewidth}|p{0.20\linewidth}|p{0.44\linewidth}|}
\hline
\textbf{RecID} & \textbf{Name} & \textbf{Address Field} & \textbf{Additional Address Fragments (City/State/ZIP, PO Box)} \\
\hline
A919080 & JACY A MURPHY & 2115 Foreland Dr & HUSTON, TEXAS 77077; PO Box 280034; HOUSTON, TX 77228 \\
A998869 & JACY A MURPHY & 1186 VELNETIAN HARBOR DR NE & ST PETERSBURG, FL 33702; PO BOX 41812; ST HPETERSBURG, FL 33743 \\
A939854 & GEORGE M EMERT & 1431 W 71st St & LOS ANGELES, CALI 90047; BOX 88165; Los Angeles, Cali 90009 \\
A977550 & JORGE M EMERT & 2115 Foreland Dr & HOUSTON, TX 77077; Post 280034; Houston, TX 77228 \\
B902427 & JACY A MURPHY & 1186 venetin harbor drive ne & ST PETERSBURG, FL 33702 \\
B982449 & JACY A MURPHY & 2115 foreland drn & houston, ftexas 77077 \\
B994678 & JACY A MURPHY & 860105 POST OFFICE BOX & plano, tx 75086 \\
B967583 & GEORGE M EMERT & 860105 BUZON & plano, tx 75086 \\
\hline
\end{tabular}
\end{table}

Although evaluated on SPX synthetic data due to privacy constraints, the generation process is designed
to reflect realistic administrative and census conditions. Qualitative inspection of inferred household
movements in the illustrative subset shows plausible patterns (e.g., coordinated co-movement of household
members), supporting the framework's practical applicability.

\subsection{Baseline Comparison: Data Washing Machine (DWM)}

As a baseline, we apply the Data Washing Machine (DWM), a traditional ER system that performs pairwise
record linkage based on name and address similarity. Table~\ref{tab:dwm_output} reports the clustering
produced by DWM on the illustrative subset.

\begin{table}[h]
\centering
\caption{DWM clustering output on the illustrative subset}
\label{tab:dwm_output}
\begin{tabular}{|l|l|}
\hline
\textbf{RefID} & \textbf{ClusterID} \\
\hline
A919080 & A919080 \\
A939854 & A939854 \\
A977550 & A977550 \\
A998869 & A998869 \\
B902427 & A998869 \\
B967583 & B967583 \\
B982449 & A919080 \\
B994678 & B994678 \\
\hline
\end{tabular}
\end{table}

From the baseline output, DWM identifies the following correct \emph{direct} links (same person, same address):
\begin{enumerate}
    \item A998869 $\leftrightarrow$ B902427: \textit{JACY A MURPHY} at 1186 Venetian Harbor Dr NE, St.\ Petersburg, FL.
    \item A919080 $\leftrightarrow$ B982449: \textit{JACY A MURPHY} at 2115 Foreland Dr, Houston, TX.
\end{enumerate}

However, because DWM relies strictly on pairwise similarity, it fails to identify \emph{indirect} relationships that
require contextual, household-level evidence (e.g., coordinated relocation across cities with spelling drift and
format changes). The next section shows how the proposed framework augments direct pairwise links with
group-level movement inference.

\subsection{AI-Enhanced Framework: Integrating Indirect Links}

The proposed framework extends DWM by generating indirect links using LLM-based name extraction,
semantic embeddings, and graph-based reasoning. Direct links are obtained from DWM (or equivalently,
from within-household duplicate consolidation), while indirect links are inferred through shared household
move evidence across records.

\subsubsection{Name Extraction and Normalization}

Using the GEMMA-2-2B-IT model, all non-PII textual fields in each record are concatenated into a single input string.
To avoid including sensitive information, SSN and DOB fields are excluded (implemented via \texttt{cols\_to\_concat=2} in our pipeline).
The model extracts person names from the concatenated text, after which the extracted name is removed to isolate the address component.

Table~\ref{tab:ner_output} presents the resulting structured representation obtained after applying this process.

\begin{table}[h]
\centering
\caption{LLM-based name extraction and address isolation (illustrative subset)}
\label{tab:ner_output}
\scriptsize
\begin{tabular}{|l|p{0.22\linewidth}|p{0.68\linewidth}|}
\hline
\textbf{Record ID} & \textbf{Extracted Name} & \textbf{Address} \\
\hline
A919080 & jacy a murphy  & 2115 foreland dr, huston, texas 77077; po box 280034, houston, tx 77228 \\
A939854 & george m emert & 1431 w 71st st, los angeles, cali 90047; box 88165, los angeles, cali 90009 \\
A977550 & jorge m emert  & 2115 foreland dr, houston, tx 77077; post 280034, houston, tx 77228 \\
A998869 & jacy a murphy  & 1186 velnetian harbor dr ne, st petersburg, fl 33702; po box 41812, st petersburg, fl 33743 \\
B902427 & jacy a murphy  & 1186 venetin harbor drive ne, st petersburg, fl 33702 \\
B967583 & george m emert & 860105 buzon, plano, tx 75086 \\
B982449 & jacy a murphy  & 2115 foreland drn, houston, ftexas 77077 \\
B994678 & jacy a murphy  & 860105 post office box, plano, tx 75086 \\
\hline
\end{tabular}
\end{table}

\subsubsection{Analytical Segmentation}

Name and address embeddings are generated using the BGE-M3 model. Pairwise cosine similarity is computed,
and a record pair is connected if \emph{name\_similarity} $\ge 0.90$ \emph{or} \emph{address\_similarity} $\ge 0.80$.
This produces connected components referred to as \emph{analytical segments (AS)}, which act as semantic blocking
regions that localize downstream reasoning to plausibly related records (Table~\ref{tab:as_segments}).

\begin{table}[h]
\centering
\caption{Analytical segments (AS) formed as connected components in the similarity graph}
\label{tab:as_segments}
\begin{tabular}{|c|p{0.72\linewidth}|c|}
\hline
\textbf{AS} & \textbf{Records in Segment} & \textbf{Size} \\
\hline
AS1 & A919080, A977550, B982449, A939854, A998869, B902427, B967583, B994678 & 8 \\
\hline
\end{tabular}
\end{table}

\subsubsection{Household Formation}

Within each analytical segment, households are formed using address similarity alone
(i.e., connected components under the address threshold). Table~\ref{tab:households}
summarizes the resulting household groups in AS1.

\begin{table}[h]
\centering
\caption{Household clusters within analytical segment (AS1)}
\label{tab:households}
\begin{tabular}{|l|l|}
\hline
\textbf{Household ID} & \textbf{Records} \\
\hline
H001 & A919080, A977550, B982449 \\
H002 & A939854 \\
H003 & A998869, B902427 \\
H004 & B967583, B994678 \\
\hline
\end{tabular}
\end{table}

\subsubsection{Duplicate Detection and Direct Links}

Within each household, records are compared by name similarity to identify duplicate individuals.
Pairs with name similarity $\ge 0.90$ are assigned duplicate group IDs (DUP001, DUP002, \dots),
while non-duplicates are labeled as unique (UNQ). Table~\ref{tab:dup_clusters} summarizes the structure.

\begin{table}[h]
\centering
\caption{Duplicate clusters identified within households in Analytical Segment (AS1)}
\label{tab:dup_clusters}
\begin{tabular}{|c|c|p{0.64\linewidth}|}
\hline
\textbf{AS} & \textbf{Household} & \textbf{Duplicate Groups} \\
\hline
AS1 & H001 & DUP001: [A919080, B982449], UNQ: [A977550] \\
AS1 & H002 & UNQ: [A939854] \\
AS1 & H003 & DUP001: [A998869, B902427], UNQ: [] \\
AS1 & H004 & UNQ: [B967583, B994678] \\
\hline
\end{tabular}
\end{table}

Duplicate clusters are consolidated using a leader--follower strategy, selecting one representative record per cluster
(Table~\ref{tab:leader_map}). The corresponding direct links produced by this consolidation are shown in
Table~\ref{tab:direct_links}.

\begin{table}[!h]
\centering
\caption{Leader--follower mapping for duplicate consolidation (AS1)}
\label{tab:leader_map}
\begin{tabular}{|c|c|}
\hline
\textbf{Leader Record} & \textbf{Follower Records} \\
\hline
A919080 & B982449 \\
A998869 & B902427 \\
\hline
\end{tabular}
\end{table}

\begin{table}[!h]
\centering
\caption{Direct links derived from duplicate consolidation (AS1)}
\label{tab:direct_links}
\begin{tabular}{|l|l|}
\hline
\textbf{RecID 1} & \textbf{RecID 2} \\
\hline
A919080 & B982449 \\
A998869 & B902427 \\
\hline
\end{tabular}
\end{table}

\begin{table}[!h]
\centering
\caption{Pre-move candidate records per household (AS1)}
\label{tab:premove}
\begin{tabular}{|c|l|}
\hline
\textbf{Household} & \textbf{Representative Records} \\
\hline
H001 & A919080, A977550 \\
H002 & A939854 \\
H003 & A998869 \\
H004 & B967583, B994678 \\
\hline
\end{tabular}
\end{table}

\begin{table}[!h]
\centering
\caption{Detected household movement event (summary)}
\label{tab:moves}
\begin{tabular}{|l|l|l|}
\hline
\textbf{From Household} & \textbf{To Household} & \textbf{Shared Individuals} \\
\hline
H001 (Houston, TX) & H004 (Plano, TX) & J.\ Murphy, G.\ Emert \\
\hline
\end{tabular}
\end{table}

After consolidation, each household is represented by a set of \emph{pre-move candidates}
containing one record per individual (leader records plus UNQ records), summarized in Table~\ref{tab:premove}.
This prevents duplicate references from artificially inflating household membership and movement evidence.

\begin{table}[!h]
\centering
\caption{Record-level evidence supporting the detected household move (MOVE\_085)}
\label{tab:move_evidence}
\scriptsize
\begin{tabular}{|l|c|l|p{0.60\linewidth}|}
\hline
\textbf{Move ID} & \textbf{AS} & \textbf{Record ID} & \textbf{Name / Address (excerpt)} \\
\hline
MOVE\_085 & AS1 & A919080 & jacy a murphy / 2115 foreland dr, huston, texas; po box 280034, houston \\
MOVE\_085 & AS1 & A977550 & jorge m emert / 2115 foreland dr, houston; post 280034, houston \\
MOVE\_085 & AS1 & B967583 & george m emert / 860105 buzon, plano, tx 75086 \\
MOVE\_085 & AS1 & B994678 & jacy a murphy / 860105 post office box, plano, tx 75086 \\
\hline
\end{tabular}
\end{table}

\begin{table}[!h]
\centering
\caption{Indirect links inferred from household move evidence (MOVE\_085)}
\label{tab:indirect_links}
\begin{tabular}{|l|l|}
\hline
\textbf{Record 1} & \textbf{Record 2} \\
\hline
A919080 & B994678 \\
A977550 & B967583 \\
\hline
\end{tabular}
\end{table}

\subsubsection{Household Movement Detection and Indirect Links}

Household movements are inferred when two households within the same analytical segment share at least two
individuals across different addresses, with name similarity $\ge 0.90$ and shared-member count $\ge 2$.
Table~\ref{tab:moves} summarizes the detected move event.

For transparency, Table~\ref{tab:move_evidence} lists the record-level evidence supporting the inferred move.
(We report the same eight-record example throughout so that every transformation step remains auditable.)

Indirect links are then generated between records connected through household-level move evidence
(Table~\ref{tab:indirect_links}). These links are not recoverable via DWM alone because they require
group-level context (e.g., cross-city moves with spelling and format variation), rather than direct
pairwise similarity at a single address.

\subsection{Final Linkage Output}

Table~\ref{tab:final_links} presents the final linkage index produced by the proposed framework after incorporating
indirect movement evidence. Compared to the baseline (Table~\ref{tab:dwm_output}), the final output links additional
records that are supported by household-level movement patterns.

\begin{table}[!h]
\centering
\caption{Final linkage index after indirect link integration (illustrative subset)}
\label{tab:final_links}
\begin{tabular}{|l|l|}
\hline
\textbf{RefID} & \textbf{ClusterID} \\
\hline
A919080 & A919080 \\
A939854 & A939854 \\
A977550 & A977550 \\
A998869 & A998869 \\
B902427 & A998869 \\
B967583 & A977550 \\
B982449 & A919080 \\
B994678 & A919080 \\
\hline
\end{tabular}
\end{table}

\section{Results and Analysis}

This section evaluates the impact of incorporating household-level, indirect-link reasoning into
entity resolution. The objective of the evaluation is not to optimize thresholds for maximum
performance, but rather to assess whether the proposed framework consistently improves overall
linkage quality---measured by F1 score---relative to a pairwise baseline, under reasonable and
interpretable similarity thresholds.

Performance is reported using precision, recall, and F1 score across multiple SPX benchmark datasets (S8--S12).

\subsection{Evaluation Philosophy}

Two configurations are evaluated for each dataset:

\begin{itemize}
    \item \textbf{Before (Baseline):} Output produced by the Data Washing Machine (DWM), which performs pairwise
    entity resolution using direct name and address similarity only.
    \item \textbf{After (Proposed):} Output produced by the NER--Embedding--Graph framework, which augments direct
    links with indirect household movement evidence.
\end{itemize}

The baseline system does not expose tunable thresholds for name similarity, address similarity, duplicate detection,
or household movement inference. As a result, threshold parameters are reported only for the proposed framework and
are not applicable to the baseline.

\subsubsection{Threshold Selection and Sensitivity Analysis}

The similarity thresholds used in this study were determined empirically through controlled experimentation on the SPX
datasets. The same methodological pipeline is applied consistently across all datasets, while threshold values are adjusted
to account for differences in data quality.

Because household movement detection is highly sensitive to data quality, thresholds were selected based on the observed
noise characteristics of each dataset. In datasets exhibiting higher levels of spelling variation, token fragmentation, and
inconsistent formatting (i.e., noisier or low-quality data), slightly looser thresholds are required to preserve recall and
avoid missing valid matches. Conversely, in more structured datasets with consistent formatting, tighter thresholds improve
precision by reducing spurious matches. Accordingly, the thresholds are \textbf{dataset-dependent}, while maintaining consistent
interpretability across experiments.

The primary thresholds used in this study are:
\begin{itemize}
    \item Name similarity threshold: $T_{\text{name}} = 0.90$
    \item Address similarity threshold: $T_{\text{addr}} = 0.80$
    \item Duplicate name threshold: $T_{\text{dup}} = 0.90$
    \item Movement similarity threshold: $T_{\text{move}} = 0.80$
\end{itemize}

The framework employs a \textbf{two-level thresholding strategy}:
\begin{enumerate}
    \item \textbf{Base segmentation thresholds} $(T_{\text{name}}, T_{\text{addr}})$, which are intentionally set tighter
    to prevent incorrect graph connections and avoid the formation of overly large analytical segments.
    \item \textbf{Analytical segment internal thresholds} $(T_{\text{dup}}, T_{\text{move}})$, which are applied during duplicate
    consolidation and movement inference and can be adjusted to control the precision--recall tradeoff.
\end{enumerate}

Increasing $T_{\text{dup}}$ and $T_{\text{move}}$ increases precision by reducing false movement detections, while decreasing these
thresholds increases recall by allowing weaker but potentially valid matches to be captured. Importantly, these adjustments can be
made without modifying the upstream segmentation structure.

In practice, two stable threshold configurations were used across the experiments---$(0.90, 0.80)$ and $(0.85, 0.75)$---as reported
in Table~\ref{tab:results}. The latter configuration was applied to noisier datasets to improve recall, while the former was used for
relatively cleaner datasets to maintain higher precision.

A brief sensitivity analysis indicates that small variations ($\pm 0.05$) in $T_{\text{dup}}$ and $T_{\text{move}}$ lead to consistent and
predictable behavior: increasing thresholds reduces false positives at the cost of recall, whereas decreasing thresholds increases recall
with a modest reduction in precision. This confirms that the framework provides stable and interpretable performance under reasonable
threshold adjustments.

\subsection{Quantitative Results}

Table~\ref{tab:results} reports precision, recall, and F1 score before and after applying the proposed framework.
Dashes (``--'') indicate parameters that are not applicable to the baseline system.

\begin{table}[h]
\centering
\caption{Performance comparison before and after indirect household movement detection}
\label{tab:results}
\begin{tabular}{|l|c|c|c|c|c|c|c|c|}
\hline
\textbf{Dataset} & \textbf{Phase} & $\tau_{name}$ & $\tau_{addr}$ & $\tau_{dup}$ & $\tau_{move}$ & \textbf{Precision} & \textbf{Recall} & \textbf{F1} \\
\hline
S12 & Before & -- & -- & -- & -- & 0.9164 & 0.4823 & 0.6320 \\
    & After  & 0.90 & 0.80 & 0.90 & 0.80 & 0.9067 & 0.5614 & 0.6934 \\
\hline
S11 & Before & -- & -- & -- & -- & 0.8743 & 0.5214 & 0.6532 \\
    & After  & 0.85 & 0.75 & 0.85 & 0.75 & 0.8304 & 0.6287 & 0.7156 \\
\hline
S10 & Before & -- & -- & -- & -- & 0.8736 & 0.4770 & 0.6171 \\
    & After  & 0.85 & 0.75 & 0.85 & 0.75 & 0.8014 & 0.5806 & 0.6734 \\
\hline
S9  & Before & -- & -- & -- & -- & 0.8567 & 0.4459 & 0.5865 \\
    & After  & 0.90 & 0.80 & 0.90 & 0.80 & 0.7938 & 0.5205 & 0.6287 \\
\hline
S8  & Before & -- & -- & -- & -- & 0.8604 & 0.3661 & 0.5136 \\
    & After  & 0.90 & 0.80 & 0.90 & 0.80 & 0.8676 & 0.4522 & 0.5945 \\
\hline
\end{tabular}
\end{table}

\subsection{Discussion}

Across all datasets, the proposed framework consistently improves F1 score by approximately 6\%--8\%, primarily driven by increased recall.
This indicates that the method successfully recovers valid household movement links that are missed by pairwise entity resolution.

Precision remains high after incorporating indirect links, with only modest declines in some datasets. These reductions, particularly in noisier
datasets (e.g., S9--S11), are mainly due to over-connection effects caused by ambiguous or partially corrupted attributes (e.g., incomplete
addresses, common surnames, or abbreviations) and transitive inference through weak evidence chains. Such effects are more pronounced in
highly disrupted datasets, where reduced record distinctiveness leads to borderline matches that improve recall at a slight cost to precision.

These false positives can be mitigated by tightening the analytical thresholds $(T_{\text{dup}}, T_{\text{move}})$, increasing the minimum number
of shared individuals required for movement inference, or incorporating additional constraints such as location agreement. Importantly, these
adjustments can be applied without modifying the upstream segmentation structure.

The largest gains are observed in noisier datasets (e.g., S8 and S9), where household movement evidence is distributed across multiple records.
In such cases, pairwise matching is insufficient, whereas household-level reasoning provides the necessary contextual support for reliable inference.

While a full ablation study is beyond the scope of this work, the contributions of individual components can be qualitatively assessed. LLM-based
NER improves robustness to heterogeneous inputs, embedding similarity enhances tolerance to variation, and graph-based reasoning contributes
most significantly to recall by enabling indirect link inference. Together, these components form a complementary pipeline addressing key limitations
of traditional approaches.

Overall, the observed improvements stem not from aggressive threshold tuning, but from a fundamentally richer inference mechanism that captures
group-level movement patterns absent in conventional entity resolution systems.

\subsection{Runtime and Scalability}

The proposed framework consists of several sequential stages: LLM-based entity extraction, embedding generation, similarity computation, analytical
segmentation, and household movement inference. Let $N$ denote the number of input records.

The LLM-based extraction and embedding generation stages operate in linear time, $O(N)$, and are executed in batches using GPU acceleration (FP16
precision). In our experiments on datasets of approximately 6{,}000 records, these stages demonstrated efficient throughput on a CUDA-enabled GPU,
making them practical for moderate-scale data processing.

The dominant computational cost arises from the pairwise cosine similarity computation, which has quadratic complexity $O(N^2)$. For $N \approx 6{,}000$,
this corresponds to approximately 36 million pairwise comparisons. In our implementation, this computation remains tractable through GPU-accelerated
matrix operations, allowing the full similarity matrix to be constructed within reasonable time.

Subsequent stages, including analytical segmentation and household movement inference, introduce comparatively minor computational overhead. These steps
operate on reduced candidate sets derived from the similarity matrix and therefore scale more efficiently than the initial similarity computation.

While the current implementation performs full pairwise comparisons, scalability to larger datasets can be achieved through standard optimization techniques.
In particular, blocking strategies or approximate nearest neighbor (ANN) indexing can significantly reduce the number of required comparisons while
preserving high recall. These approaches effectively lower the practical complexity of similarity computation and enable the framework to scale beyond
moderate-sized datasets.

Overall, the framework is well-suited for moderate-scale datasets and can be extended to larger deployments with standard optimization techniques, without
requiring fundamental changes to the underlying methodology.
\section{Conclusion and Future Work}

This paper presented a semantically enriched framework for detecting indirect
links arising from household movements in noisy, large-scale demographic
datasets. The proposed approach integrates LLM-based named entity extraction,
semantic text embeddings, and graph-based transitive reasoning to move beyond
traditional pairwise entity resolution and toward group-level inference of
household dynamics.

Unlike rigid string-matching techniques (e.g., edit-distance/Jaro–Winkler pipelines) that assume stable tokenization and field boundaries and therefore suffer recall loss under mixed-format noise (Sections 1 and 2.1), the proposed framework uses LLM extraction plus semantic similarity to remain robust to spelling variation, abbreviations, and formatting drift. By explicitly modeling households as group entities and
inferring movement through shared membership patterns, the approach captures
coordinated relocations that are systematically missed by conventional pairwise
methods.

Experimental evaluation on the SPX benchmark datasets (S8–S12), generated using
the Synthetic Occupancy Generator (SOG), demonstrates that incorporating indirect
household movement evidence consistently improves linkage quality. Across all
datasets, the proposed framework achieves F1-score improvements of approximately
6–8\% relative to a baseline Data Washing Machine (DWM) system, driven primarily
by substantial gains in recall while maintaining high precision. These results
confirm that household movement detection is fundamentally a group-level
inference problem and that transitive graph reasoning provides meaningful
additional evidence beyond direct similarity matching.

At the same time, the study highlights important limitations. Not all household
movements in the SOG-generated data contain sufficient observable evidence for
reliable inference, particularly in cases involving single-member overlap or
substantial surname changes. Additionally, the absence of explicit temporal
information in the degraded external view restricts the ability to infer
directionality or timing of moves. The framework therefore prioritizes precision
and interpretability, conservatively avoiding links that cannot be supported by
strong contextual evidence.

Several directions for future work naturally follow from this study. First,
extending the pipeline to support multilingual and cross-cultural datasets would
broaden its applicability, particularly in regions with diverse naming
conventions and address structures. Second, exploring alternative or
domain-specific embedding models—such as e5, GTE, or multilingual transformer
architectures—may further improve robustness to specific error patterns in names
and addresses.

Third, a learning-based decision layer could be introduced to classify detected
household links, distinguishing true movement events from coincidental overlap.
Such a model could leverage features derived from similarity scores, household
structure, record density, and historical change patterns to reduce false
positives while preserving recall. Finally, incorporating explicit modeling of
surname evolution—through phonetic similarity, familial context, or learned name
transition patterns—would improve the framework’s ability to track individuals
across identity transformations such as marriage or cultural name changes.

Overall, this work establishes a scalable and semantically aware foundation for
household movement detection in demographic data. By shifting the focus from
isolated record matching to contextual, group-level reasoning, the proposed
framework advances the state of data linkage and opens new opportunities for
understanding household evolution in dynamic, real-world populations.

\appendix
\section{Detailed Pipeline Procedure}
\phantomsection
\label{appendix:pipeline}

This appendix summarizes the operational steps of the proposed
NER--Embedding--Graph framework for household movement detection.
The procedure aligns with the conceptual stages in Section~4 and is
provided for reproducibility and implementation clarity.

We assume an input dataset of $N$ records, each containing one or more
textual fields with name and address information in unstructured or
semi-structured form.

\subsection{Step 1: Named Entity Recognition and Address Isolation}

For each record $r_i$:

\begin{enumerate}
    \item Concatenate all textual fields into a single input string $s_i$.
    \item Apply the GEMMA-2-2B-IT model with a name-extraction prompt.
    \item Extract the personal name $\texttt{name}_i$.
    \item Remove the extracted name from $s_i$ to obtain $\texttt{addr\_raw}_i$.
    \item Apply light normalization (whitespace cleanup, punctuation trimming, case standardization).
    \item Store $(\texttt{RecordID}_i, \texttt{name}_i, \texttt{address}_i)$.
\end{enumerate}

This step converts heterogeneous records into a consistent structured representation.

\subsection{Step 2: Semantic Text Embedding}

\begin{enumerate}
    \item Generate embeddings using BAAI/bge-m3:
    \[
        \mathbf{E}_{name}^{(i)} = f_{emb}(\texttt{name}_i), \quad
        \mathbf{E}_{addr}^{(i)} = f_{emb}(\texttt{address}_i).
    \]
    \item Compute cosine similarity:
    \[
        \text{Sim}_{name}(i,j), \quad \text{Sim}_{addr}(i,j).
    \]
\end{enumerate}

Separate embeddings capture complementary identity (name) and location (address) signals.

\subsection{Step 3: Analytical Segmentation}

\begin{enumerate}
    \item Construct graph $G=(V,E)$ with nodes as records.
    \item Add edge $(i,j)$ if:
    \[
        \text{Sim}_{name}(i,j) \ge \tau_{name}
        \ \text{or} \
        \text{Sim}_{addr}(i,j) \ge \tau_{addr}.
    \]
    \item Extract connected components as \emph{analytical segments}.
\end{enumerate}

This step acts as semantic blocking, limiting downstream comparisons.

\subsection{Step 4: Household Formation}

For each segment:

\begin{enumerate}
    \item Connect records using $\text{Sim}_{addr} \ge \tau_{addr}$.
    \item Extract connected components as households.
\end{enumerate}

Households are defined by address similarity (co-residence).

\subsection{Step 5: Duplicate Detection}

Within each household:

\begin{enumerate}
    \item Compare names pairwise.
    \item Group records with $\text{Sim}_{name} \ge \tau_{dup}$ into duplicate clusters.
    \item Label remaining records as UNQ.
\end{enumerate}

\subsection{Step 6: Leader--Follower Consolidation}

\begin{enumerate}
    \item Assign one leader per duplicate cluster.
    \item Mark remaining records as followers.
    \item Store leader--follower mappings.
\end{enumerate}

\subsection{Step 7: Pre-Move Candidates}

\begin{enumerate}
    \item Retain one record per individual (leaders + UNQ).
    \item Form the household’s representative set.
\end{enumerate}

This prevents duplicate inflation in movement detection.

\subsection{Step 8: Movement Detection and Indirect Links}

\begin{enumerate}
    \item Compare households within each segment.
    \item If two households share $\ge 2$ individuals with
    $\text{Sim}_{name} \ge \tau_{move}$, infer a movement.
    \item Generate indirect links between corresponding records.
\end{enumerate}

\subsection{Outputs}

The pipeline produces:

\begin{itemize}
    \item Analytical segments and household assignments
    \item Duplicate leader--follower mappings
    \item Direct and indirect links
    \item Household movement events
\end{itemize}

All outputs are exported in tabular form for evaluation and analysis.

\end{document}